\documentclass{article}

\PassOptionsToPackage{numbers, sort&compress}{natbib}
\usepackage[final]{nips_2016}

\usepackage[utf8]{inputenc}       
\usepackage[T1]{fontenc}          
\usepackage{hyperref}             
\usepackage{url}                  
\usepackage{booktabs}             
\usepackage{amsfonts}             
\usepackage{nicefrac}             
\usepackage{microtype}            
\usepackage{amssymb}              
\usepackage{mathtools}            
\usepackage{mleftright}           
\usepackage{graphicx}             
\usepackage{tikz}                 
\usepackage{pgfplots}             
\usepackage{paralist}             
\usepackage[capitalise]{cleveref} 
\usepackage{textcomp}
\usepackage{wrapfig}

\graphicspath{{images/}}
\usetikzlibrary{graphs, bayesnet, shapes, shapes.symbols, patterns}
\tikzset{const/.append style={inner sep=2pt}}
\tikzset{det/.style={const,draw,minimum size=18pt,node distance=0.6}}
\tikzset{loss/.style={det,fill=grey,text=white}}
\tikzset{sto/.style={circle,draw,minimum size=18pt,node distance=0.5}}
\tikzset{>={stealth}}

\pgfdeclarepatternformonly{hash bars}{\pgfqpoint{-1pt}{-1pt}}{\pgfqpoint{4pt}{4pt}}{\pgfqpoint{3pt}{3pt}}%
{
  \pgfsetlinewidth{0.9pt}
  \pgfpathmoveto{\pgfqpoint{0pt}{3pt}}
  \pgfpathlineto{\pgfqpoint{3.1pt}{-0.1pt}}
  \pgfusepath{stroke}
}
\tikzstyle{partial-obs} = [circle,pattern=hash bars,pattern color=gray!25,draw=black,inner sep=1pt,
minimum size=20pt, font=\fontsize{10}{10}\selectfont, node distance=1]
\tikzstyle{plate caption} = [caption, node distance=0, inner sep=0pt,
below left=0em and 0em of #1.south east] %

\newcommand{\defoccur}[1]{\textsl{#1}}

\newcommand{\ppm}{\ensuremath{\pm}\,}

\usepackage{scalerel}

\DeclareMathOperator{\E}{\mathbb{E}}
\DeclareMathOperator{\Lbo}{\mathcal{L}}

\renewcommand{\to}{\ensuremath{\rightarrow}}

\newcommand{\given}{\mid}

\newcommand{\fnP}[2]{\ensuremath{#1 \mleft( #2 \mright)}}

\newcommand{\fnS}[2]{\ensuremath{#1 \mleft[ #2 \mright]}}

\newcommand{\ex}[2]{\fnS{\E_{#1}}{#2}}
\newcommand{\KL}[2]{\fnP{D_{\text{\scalebox{0.75}{KL}}}}{#1 \,\|\; #2}}
\newcommand{\p}[2][]{\fnP{p_{#1}}{#2}}
\newcommand{\q}[2][]{\fnP{q_{#1}}{#2}}

\newcommand{\data}[1][]{\ensuremath{\mathbf{x}^{#1}}}

\newcommand{\latent}[1][]{\ensuremath{\mathbf{z}^{#1}}}

\newcommand{\llabel}[1][]{\ensuremath{\mathbf{y}^{#1}}}

\newcommand{\varP}{\ensuremath{\phi}}
\newcommand{\modelP}{\ensuremath{\theta}}

\newcommand{\modelF}[1]{\p[\modelP]{#1}}
\newcommand{\modelCF}[2]{\p[\modelP]{#1 \given #2}}
\newcommand{\varF}[1]{\q[\varP]{#1}}
\newcommand{\modelGenT}{\modelF{\data, \latent}}                   
\newcommand{\modelPriorT}[1][i]{\p{\latent}}                       
\newcommand{\modelLikT}[1][i]{\modelCF{\data}{\latent}}            
\newcommand{\modelPosT}{\modelCF{\latent}{\data}}                  
\newcommand{\varPosT}{\varF{\latent \given \data}}                 
\newcommand{\varEx}[1]{\ex{\varPosT}{#1}}

\newcommand{\elboT}{\fnP{\Lbo}{\modelP,\varP; \data}}

\title{Inducing Interpretable Representations with Variational Autoencoders}

\author{
  N. Siddharth \\
  University of Oxford\\
  \texttt{nsid@robots.ox.ac.uk} \\
  \And
  Brooks Paige \\
  University of Oxford\\
  \texttt{brooks@robots.ox.ac.uk} \\
  \And
  Alban Desmaison \\
  University of Oxford\\
  \texttt{alban@robots.ox.ac.uk} \\
  \And
  Jan-Willem Van de Meent \\
  Northeastern University\\
  \texttt{j.vandemeent@northeastern.edu} \\
  \And
  Frank Wood \\
  University of Oxford\\
  \texttt{fwood@robots.ox.ac.uk} \\
  \And
  Noah D. Goodman \\
  Stanford University\\
  \texttt{ngoodman@stanford.edu} \\
  \And
  Pushmeet Kohli \\
  Microsoft Research\\
  \texttt{pkohli@microsoft.com} \\
  \And
  Philip H.S. Torr \\
  University of Oxford\\
  \texttt{philip.torr@eng.ox.ac.uk} \\
}

\begin{document}
\maketitle

\begin{abstract}
  We develop a framework for incorporating structured graphical models in the
  \emph{encoders} of variational autoencoders (VAEs) that allows us to induce
  interpretable representations through approximate variational inference.
  This allows us to both perform reasoning (e.g. classification) under the
  structural constraints of a given graphical model, and use deep generative
  models to deal with messy, high-dimensional domains where it is often
  difficult to model all the variation.
  Learning in this framework is carried out end-to-end with a variational
  objective, applying to both unsupervised and semi-supervised schemes.
\end{abstract}

\section{Introduction}
\label{sec:intro}

Reasoning in complex perceptual domains such as vision often involves two
facets: the ability to effectively learn flexible representations of the complex
high-dimensional data, and the ability to interpret the representations in some
structured form.
The former is a measure of how well one can capture the relevant information in
the data, and the latter is a means of employing consistent semantics to such,
in an effort to help diagnosis, enable composition, and improve generality.

Probabilistic graphical models\cite{koller2009probabilistic,murphy2012machine}
enable structured representations, but often in perceptual domains such as
vision, they require extensive specification and significant feature engineering
to be useful.
Variational Autoencoders (VAEs) \cite{kingma2013auto,rezende2014stochastic}, are
a form of generative model, where the (typically) manually specified feature
extractors are replaced with (deep) neural networks.
Here, parameters of both the generative model and an approximation to the true
posterior, called the \emph{recognition} model, are learned simultaneously.
However, a particular feature of such approximations is that they exhibit
entangled, and non-interpretable, latent representations by virtue of the fact
that the approximating distributions are assumed to take a general, flexible
form; typically multivariate normal.

Our contribution extends the combination of deep neural networks and graphical
models to allow the use of \emph{arbitrarily structured} graphical models as
variational approximations, which enforces latent representations to conform to
the types and structure of the provided graphical model.
And where the structure alone is insufficient to encourage disentangled
representations, we further extend this framework to perform semi-supervised
learning, using a handful of labelled data to help disentangle the latent
representation.\footnote{
  For the purposes of this manuscript, we refer to latent representations that
  are disentangled as \emph{structured} and latent representations that are
  entangled as \emph{unstructured}.
  The notions of entangled and disentangled representations relate to concise
  and well-defined human interpretability (visual gestalt) of the axes of
  variation.
}
Our framework employs a single variational objective in which parameters of both
the generative and recognition models are learned simultaneously.

We shares features, motivation, and goals with a variety of recent work.
\citet{kingma2014semi} explores the ability to perform semi-supervised learning
in the VAE setting.
This is accomplished by partitioning the latent space into structured and
unstructured random variables, and providing labels for the structured
variables.
\citet{kulkarni2015deep} employ an particular interpretable model for their
latent space, where each component is independent of the others, providing weak
supervision through a customized training procedure rather than through explicit
labels.
We build on such work on semi-supervised learning by extending to more general
models and structures for the latent space.
\citet{sohn2015learning} perform fully-supervised learning in the particular
case where both the (unstructured) latents and labels can be taken to be
conditioned on the data.

Closest in spirit and motivation is recent work by \citet{svae016nips}, which
also involves combining graphical models with VAEs to do unsupervised learning.
It is employed as a means to extend the class of problems for which graphical
model inference for can be performed effectively, involving the relaxation of
conjugacy constraints for likelihoods.
Finally, \citet{schulman2015gradient} provides a general method for estimating
gradients of stochastic computations, which has been applied to models with
structured latent spaces and discrete latent variables by
\citet{AliEslami2016AIR}.
An additional contribution of our work is a package for Torch
\citep{collobert2011torch7} which permits simple simultaneous specification of
deep generative models with structured latent spaces, and of the their
corresponding inference networks.

\section{Formulation}
\label{sec:formulation}

Fundamentally, we wish to learn the parameters of a graphical model chosen to
model the data.
This is typically a generative model over data~\data and latents~\latent,
denoted~\modelGenT.
We would like to estimate the posterior over the latents given the data,
denoted~\modelPosT, in order to extract a representation.
When we wish to extract an \emph{interpretable} representation, then this
corresponds to constraining the model we are learning to be one whose posterior
distribution is then amenable to human inspection.

Although in the general case, computation of the exact posterior
distribution~\modelPosT{} is intractable, recent advances in deep generative
models enable the the use of the variational autoencoder to learn a
parametrised approximation~\varPosT{} to it.
Here, the variational approximation is used as a surrogate for the (intractable)
exact posterior, constrained to match the true posterior through
\(\KL{\varPosT}{\modelPosT}\).
However, since one cannot actually evaluate the true posterior, the VAE optimises
an alternate objective
\begin{align*}
  \KL{\varPosT}{\modelPosT} &= -\elboT + \log \modelF{\data}\\
  \text{where\,} \elboT &= \varEx{\modelGenT - \varPosT}
\end{align*}
called the \defoccur{evidence lower bound} (ELBO) that lower bounds the marginal
likelihood~\(\log \modelF{\data}\).
Here, both the generative model parameters~\modelP{} and recognition model (the
approximation distribution) parameters~\varP{} are characterised by (deep)
neural networks, and are both learned simultaneously.
The ELBO objective can also be reformulated as
\[
  \elboT = \varEx{\modelLikT} - \KL{\varPosT}{\modelPriorT}
\]
to indicate that the approximating distribution is used, along with a prior over
the latents, to \emph{regularise} the standard autoencoder objective of the
expected log likelihood.

While recent approaches to deep generative modelling places constraints, on the
structure of the generative model~\modelGenT \citep{svae016nips}, we incorporate
them into the \emph{encoder model}~\varPosT.
We do so for two principal reasons.
Firstly, a mean-field approximation in \(\varF{\latent \given \data}\), as is
typically assumed, is a poor fit for perceptual domains such as vision.
Complex dependencies that arise in the posterior due to intricacies of the
rendering process, even when latent variables may be considered \emph{a priori}
independent, means that such a mean-field assumption is often insufficient.
Secondly, an unstructured form (say, multivariate normal) for the variational
approximation, means that the recognition model produces latents that are also
unstructured, and as is, not interpretable.
Any attempts to imbue an interpretation on such representations typically
happens after the fact, by adding a discriminative model on top of the learned
representations.
Adding structure to the encoder model ameliorates both these concerns, by
allowing a richer dependency structure in the recognition model, and also
inducing latent representations whose interpretability is governed by the given
graphical model.
Our framework enables the specification of a wide variety of graphical models,
in an embedded domain-specific language (EDSL), expressed directly in the
Torch\cite{collobert2011torch7} framework.

\vspace*{-2ex}
\subsection{Model}
\label{sec:model}
Particularly, for the domains we are interested in here, the models we employ
factorise into structured latents~\llabel{} and unstructured latents~\latent, on
top of the specific factorisation imposed for the structured latent variables.
The typical form of the generative model is given by
\(
  \modelF{\data, \latent \given \llabel}
  = \modelF{\data \given \latent, \llabel} \modelF{\latent, \llabel}
\)
where \(\modelF{\data \given \latent, \llabel}\) is typically a multivariate
normal distribution and \(\modelF{\latent, \llabel}\) is some appropriately
structured latent(s).
We use the unstructured latent variables as a means to capture variation in the
data not explicitly modelled, jointly learning a likelihood function partially
constrained by the structured latents, but crucially not enforcing that they
totally explain the data.

The variational approximation to the true posterior, \varPosT{}, is nominally
taken to be of the same family as the prior distribution, as
\(\varF{\latent, \llabel \given \data}\), but can often include additional
structure and alternate factorisations as appropriate.
One particular factorisation introduces a dependence between the structured and
unstructured latents in the approximation, conditioning the latter on the former
as
\(
  \varF{\latent, \llabel \given \data}
  = \varF{\latent \given \llabel, \data} \varF{\llabel \given \data}.
\)
This removes the implicit ``mean field'' assumption in the recognition network,
and reflects the fact that the latent variables~\latent{} and~\llabel{}
typically exhibit conditional dependence on~\data, even if the latent variables
are \emph{a priori} independent.

Models with such top-level factoring are useful for situations where
interpretability is only required or useful to model along certain axes of
variation.
It is useful when we wish to interpret the same data from different viewpoints
and contexts like when the choice and form of labels is fixed.
And it is useful for when we cannot conceivable capture all the variation in the
data due to its complexity and so settle for a particular restriction, as is the
case with real world visual and language data.

\vspace*{-2ex}
\subsection{Learning}
\label{sec:learning}
Although we impose structure in the recognition network through the graphical
models, it is not necessarily certain that the nodes corresponding to particular
variables actually encode the desired ``semantics'' of that node.
For example, in a graphical model that decomposes as described above, where the
structured latent~\llabel{} encodes digit identity (0-9), and the unstructured
latent~\latent{} captures the style, there is no certainty that the
decomposition alone is sufficient to learn disentangled representations
Without the use of supervision, one has no guarantee that the structured and
unstructured latents fulfil their respective roles in such a scheme.

We build on the work by \citet{kingma2014semi} to construct a semi-supervised
learning scheme where a small amount of supervision is sufficient to break the
inherent symmetry problem and learn appropriate representation.
In their framework, the objective has a term involving labelled data, that
treats both data~\data{} and label~\llabel{} as \emph{observed} variables, and a
term involving unlabelled data, that simply \emph{marginalises out} the
label~\llabel over its support.
They also add an explicit term to learn a classifier (in the recognition model)
on the supervised data points.

We too can employ the same objective, but we note that in such cases, there is
often a cost to be paid computationally.
The marginalisation scales poorly with both shortage of labels and support size.
%
Alternately, we observe that for discrete random variables they are only used as
input to the neural network that parametrises the generative model, we can often
simply \emph{plug-in} the probability vector of the discrete distribution
instead of sampling from it, similar to the \emph{straight-through} estimator
\citep{bengio2013estimating}.
This is of course, not applicable in general, but if the posterior over
labels~\(\modelF{\llabel \given \data}\) is close to a Dirac-Delta function, as
in the classifying-digits example, then it is a good approximation.

Other points of difference involve the use of richer approximations for the
encoder and decoder in the form of \defoccur{convolutional neural networks}
(CNNs) \citep{imagenet2012nips}, and the introduction of a \emph{supervision
  rate} enabling repeated observation of a labelled data point, in different
contexts, in order to reduce estimator variance.
CNNs helps avoid employing a stacked model \citep{kingma2014semi}, allowing a
single, joint objective with comparable performance.
Supervision rates are motivated by the fact that observing a labelled data point
in the context of \emph{different} unlabelled data points (in a mini-batched
training regime), can help moderate the variance in learning steps.

\section{Experiments}
\label{sec:expts}
\begin{wrapfigure}[8]{r}{0.37\textwidth}
  \centering
  \vspace*{-3ex}
  \begin{tikzpicture}
    \node[obs](x){\data};
    \node[latent, left=0em of x,yshift=3em] (n) {${n}$};
    \node[partial-obs, right=0em of x,yshift=3em] (d) {${d}$};
    \edge {n,d} {x} ;
  \end{tikzpicture}
  \hspace{1ex}
  \begin{tikzpicture}
    \node[obs](x){\data};
    \node[latent, left=0em of x,yshift=3em] (n) {${n}$};
    \node[latent, right=0em of x,yshift=3em] (d) {${d}$};
    \edge{x}{n,d};
    \edge{d}{n};
  \end{tikzpicture}
  \caption{%
    (l) Generative and (r) recognition models with digit~\(d\) and style~\(n\).
  }
  \label{fig:gm-mnist}
\end{wrapfigure}
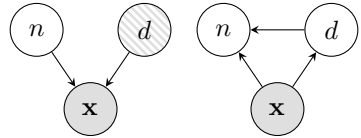
We evaluate our framework on its ability to learn interpretable latents through
both an effective recognition model and an effective generative model.
The efficacy of the recognition model is evaluated on a label-classification
task, and the efficacy the generative model is evaluated on the \emph{visual
  analogies} task.
The evaluations are conducted on both the MNIST and Google Street-View House
Numbers (SVHN) datasets using the generative and recognition models shown in
\cref{fig:gm-mnist}.

Both the MNIST and SVHN datasets were employed with a training-test split of
60000/10000 for MNIST and 73000/26000 for SVHN.
For the MNIST dataset, we use a standard single-hidden-layer MLP with 512 modes
for both the encoder and decoder.
For the SVHN dataset, we use a CNN architecture with a convolutional encoder and
a deconvolutional decoder, with two blocks of 32 \(\to\) 64 filters in the
encoder, and the reverse in the decoder.
For learning, we used AdaM \citep{KingmaB14Adam} with a learning rate of 0.001
(0.0003 for SVHN) and momentum-correction terms set to their default values.
The minibatch sizes varied from 80-300 depending on the dataset used and the
supervised-set size.

\begin{figure}
  \centering
  \begin{tabular}{@{}cc@{}}
    \scalebox{0.9}{%
    \begin{tabular}[b]{@{}c@{\hspace*{3ex}}cc@{}}
      \toprule
      \multicolumn{3}{c}{MNIST}\\
      \midrule
      \(l\) & Ours & ``M2'' \citep{kingma2014semi}\\
      \cmidrule(r){1-1} \cmidrule(r){2-3}
      10  & 12.2 (\ppm 1.38) & 11.97 (\ppm 1.71)\\
      60  & 5.28 (\ppm 0.76) & 4.94 (\ppm 0.13)\\
      100 & 4.23 (\ppm 0.68) & 3.60 (\ppm 0.56)\\
      300 & 3.94 (\ppm 0.77) & 3.92 (\ppm 0.63)\\
      \bottomrule\addlinespace[5pt]
      \toprule
      \multicolumn{3}{c}{SVHN}\\
      \midrule
      \(l\) & Ours & ``M1+M2'' \citep{kingma2014semi}\\
      \cmidrule(r){1-1} \cmidrule(r){2-3}
      100 & 30.32 (\ppm 2.74) &36.02 (\ppm 0.10)\\
      300 & 23.98 (\ppm 1.83) &-\\
      \bottomrule\addlinespace[4ex]
    \end{tabular}
    }
  &
    \includegraphics[width=0.52\textwidth]{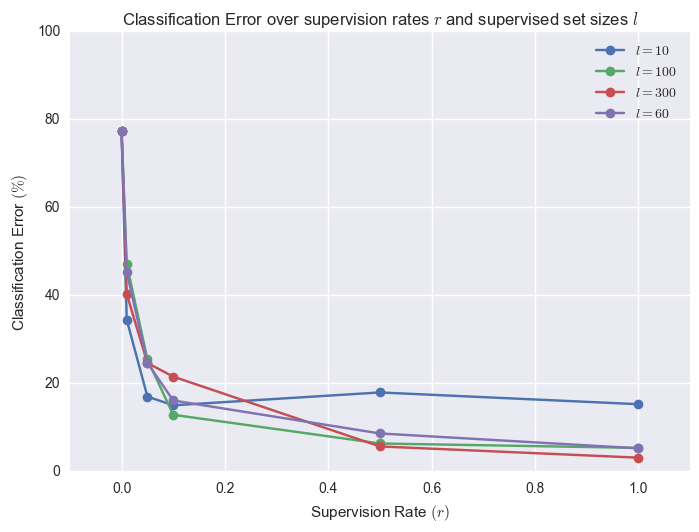}\\[-1ex]
    (a) & (b)
  \end{tabular}
  \caption{
    (a) Classification-error rates for different (per-class) labelled-set
    sizes (\(l\)) over different runs.
    (b) Classification-error for the MNIST dataset over different labelled set
    (per class) sizes (\(l\)) and supervision rates (\(r\)) = \{0, 0.01, 0.05, 0.1,
      0.5, 1.0\}.
  }
  \label{fig:error}
  \vspace*{-2ex}
\end{figure}

To evaluate the recognition model quantitatively, we compute the classification
accuracy of the label-prediction task with the model for both datasets.
This allows us to measure the extent to which the latent-space representations
are disentangled, capturing the kinds of representations one would expect
\emph{a priori} given the graphical model.
The results, with comparison against \citet{kingma2014semi}, are reported in
\cref{fig:error}(a).
For the MNIST dataset, we compare against their ``M2'' model, as we use just the
standard MLP for the experiments without performing a preliminary
feature-learning step.
For the SVHN dataset, we compare against the stacked ``M1+M2'' model, since we
employ a more effective feature learner for visual data through the CNN.
As can be seen from the results, we perform comparably on the MNIST dataset, and
comfortably beat the error rates on the SVHN dataset.
Note that these recognition networks employed the plug-in estimator discussed in
\cref{sec:learning}.

A particular feature of our approach is the ability to learn disentangled
representations with just a few labelled data points.
Combined with the ability to re-observe a particular labelled data point through
the use of the supervision rate, our framework can effectively disentangle the
latent representations in a semi-supervised learning regime involving only a
handful of labelled data.
\Cref{fig:error}(b) shows how the error rate varies with change in the
supervision rate for different labelled set (per class) sizes.
Note the steep drop in error rate with just a handful of labels (e.g.\ 10) seen
just a few times (e.g.\ 1\% of the time).
The supervision rate here corresponds to sampling minibatches of 80 data points
from a total labelled set of 100 data points, with each label class equally
represented in the labelled set.

\begin{figure}
  \centering
  \begin{tabular}{@{}cc@{\hspace*{3pt}}c@{}}
    \begin{tabular}[b]{@{}c@{}}
      \includegraphics[width=0.27\textwidth]{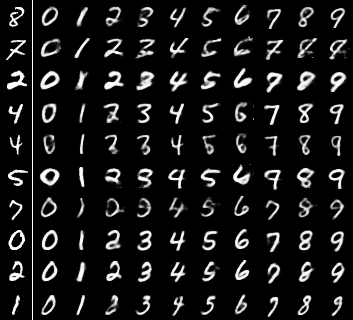}\\
      \includegraphics[width=0.27\textwidth]{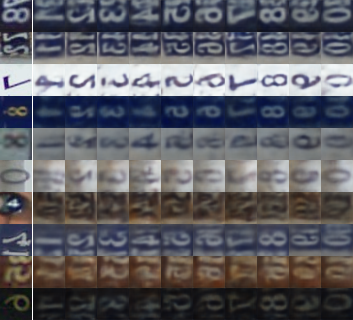}
    \end{tabular}
 &
   \includegraphics[width=0.16\textwidth]{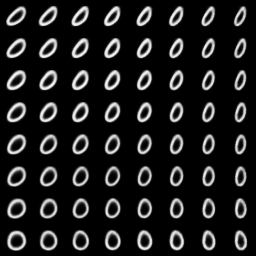}
 &
   \begin{tabular}[b]{@{}c@{\hspace*{3pt}}c@{\hspace*{3pt}}c@{}}
     \includegraphics[width=0.16\textwidth]{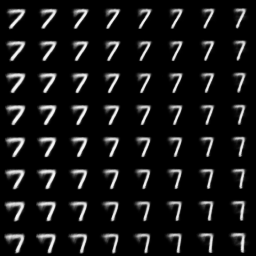}
     &\includegraphics[width=0.16\textwidth]{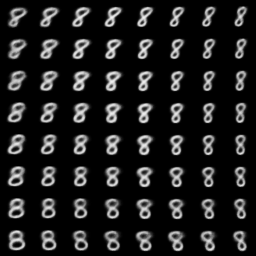}
     &\includegraphics[width=0.16\textwidth]{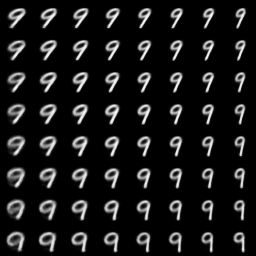}\\[-1pt]
     \includegraphics[width=0.16\textwidth]{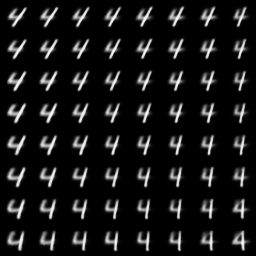}
     &\includegraphics[width=0.16\textwidth]{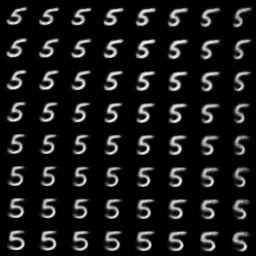}
     &\includegraphics[width=0.16\textwidth]{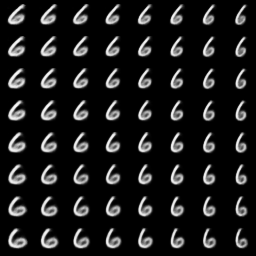}\\[-1pt]
     \includegraphics[width=0.16\textwidth]{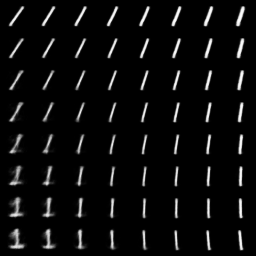}
     &\includegraphics[width=0.16\textwidth]{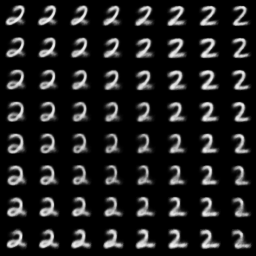}
     &\includegraphics[width=0.16\textwidth]{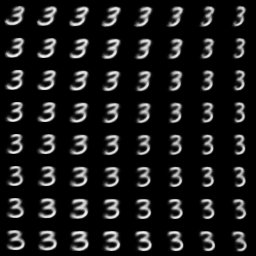}
   \end{tabular}\\
    (a)&&(b)\hspace*{10ex}\hfill
  \end{tabular}
  \caption{%
    Exploring the disentangled latent space through the generative model.
    (a) Visual analogies, where the style latent variable~\(n\) is kept fixed
    and the label~\(l\) varied.
    (b) Exploration in the style~\(n\) space for a 2D latent Gaussian random
    variable, keeping label~\(l\) fixed.
  }
  \label{fig:interpretable}
  \vspace*{-2ex}
\end{figure}

Another means of measuring how well the latent space has been disentangled is by
manipulation of the generative model.
Here, one can vary the values of particular variables, and observe if the
generative model produces outputs that suitably reflect the changes effected.
For the datasets and models considered here, this is cast as the visual
analogies task.
\Cref{fig:interpretable} demonstrates the effect of manipulating the latent
variables in the learnt generative model in different ways.

\Cref{fig:interpretable}(a) tests the changes observed in the generative model
outputs when the style variable~\(n\) is held constant, and the digit
label~\(l\) is varied.
For both the MNIST and SVHN datasets, it clearly demonstrates that changing only
the digit label has the expected effect of varying the class, but maintaining
style.
Had the latent space not been sufficiently disentangled, this could not be the
case.

\Cref{fig:interpretable}(b) tests the changes observed in the generative model
outputs in the opposite case, when the digit label~\(l\) is held constant, and
the style variable~\(l\) is varied, for each of the digits in the MNIST dataset.
Note that we only evaluate this capability on the MNIST dataset as this
particular exercise needs the style variable to be 2-dimensional, which is just
sufficient to capture the variations in MNIST, but is not sufficient to capture
variation in the more complex SVHN dataset.
Again, we note that digits maintain their identity in the outputs while
\emph{systematically} reflecting changes in style.
This also is something that would not be possible had the latents not been
sufficiently disentangled.

In summary, we demonstrate the utility and efficacy of employing graphical
models in the encoders or recognition networks of variational autoencoders to
induce interpretable latent representations with semi-supervised learning.
Results of experiments conducted with our framework demonstrate, both
qualitatively and quantitatively, the practical effectiveness of our framework
in learning interpretable and disentangled latent representations.

\bibliographystyle{plainnat}
\bibliography{citations}

\end{document}